\documentclass[letterpaper]{article} 
\usepackage{aaai25}  
\usepackage{times}  
\usepackage{helvet}  
\usepackage{courier}  
\usepackage[hyphens]{url}  
\usepackage{graphicx} 
\usepackage{natbib}  
\usepackage{caption} 
\usepackage{algorithm}
\usepackage{algorithmic}

\usepackage{newfloat}
\usepackage{listings}
\usepackage{booktabs}
\usepackage{multirow}
\DeclareCaptionStyle{ruled}{labelfont=normalfont,labelsep=colon,strut=off} 
\floatstyle{ruled}
\newfloat{listing}{tb}{lst}{}
\floatname{listing}{Listing}
\nocopyright
\title{Boosting Lossless Speculative Decoding via \\Feature Sampling and Partial Alignment Distillation}
\author {
    Lujun Gui\textsuperscript{\rm 1},
    Bin Xiao\textsuperscript{\rm 2},
    Lei Su\textsuperscript{\rm 2},
    Weipeng Chen\textsuperscript{\rm 2}
}
\affiliations {
    \textsuperscript{\rm 1}Beijing Institute of Technology \ 
    \textsuperscript{\rm 2}Baichuan Inc.\\
    lujun.gui@bit.edu.cn\\
    \{xiaobin,sulei,chenweipeng\}@baichuan-inc.com
}

\usepackage{bibentry}

\begin{document}

\maketitle

\begin{abstract}
Lossless speculative decoding accelerates target large language model (LLM) inference by employing a lightweight draft model for generating tree-structured candidates, which are subsequently verified in parallel by the target LLM.
Currently, effective approaches leverage feature-level rather than token-level autoregression within the draft model to facilitate more straightforward predictions and enhanced knowledge distillation.
In this paper, we reassess these approaches and propose FSPAD (Feature Sampling and Partial Alignment Distillation for Lossless Speculative Decoding), which introduces two straightforward and effective components within the existing framework to boost lossless speculative decoding.
Firstly, FSPAD utilizes token embeddings to sample features of the target LLM in high-dimensional space before feeding them into the draft model, due to the inherent uncertainty of the features preventing the draft model from obtaining the specific token output by the target LLM.
Secondly, FSPAD introduces partial alignment distillation to weaken the draft model's connection between features and logits, aiming to reduce the conflict between feature alignment and logit confidence during training.
Our experiments include both greedy and non-greedy decoding on the largest and smallest models from the Vicuna and LLaMA3-Instruct series, as well as tasks in multi-turn conversation, translation, summarization, question answering, mathematical reasoning, and retrieval-augmented generation.
The results show that FSPAD outperforms the state-of-the-art method across all the aforementioned tasks and target LLMs.
\end{abstract}

%

\begin{figure}[t]
\centering
\includegraphics[width=1.0\columnwidth]{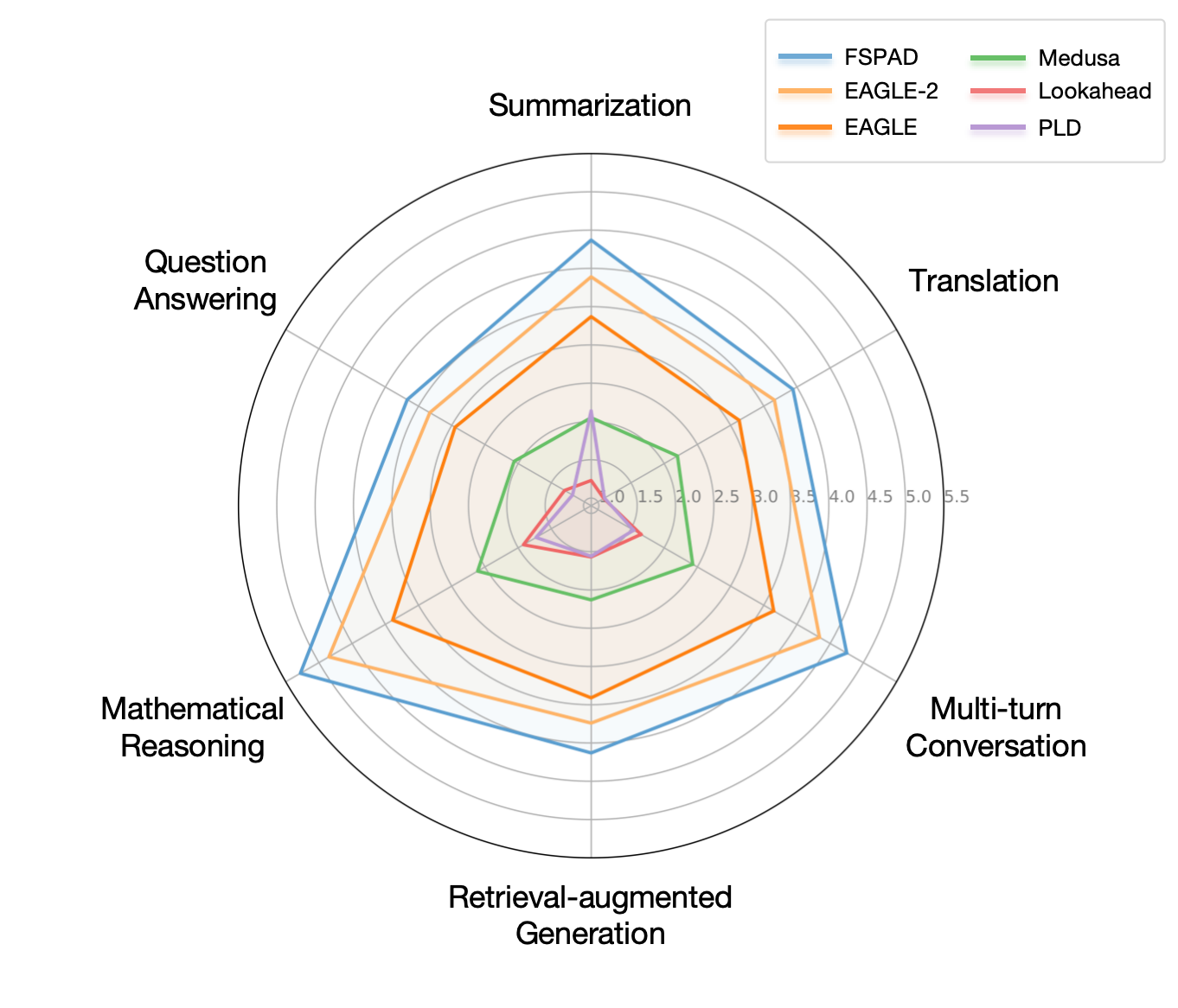} 
\caption{The number of tokens generated per step by Vicuna 33B during greedy decoding in tasks of multi-turn conversation, translation, summarization, question answering, mathematical reasoning, and retrieval-augmented generation. \textit{In this paper, we exclusively compare lossless speculative decoding methods to ensure that the distribution of the output text remains unchanged.}}
\label{page1}
\end{figure}

\section{Introduction}
Large Language Models (LLMs) \cite{achiam2023gpt} demonstrate remarkable abilities and are extensively utilized in various fields.
Autoregressive generation, the prevailing standard for LLMs, generates the next token sequentially, resulting in an expensive and slow inference process.
Lossless speculative decoding \cite{chen2023accelerating} addresses this by splitting the process of target LLMs into a low-cost draft phase and a parallel verification phase of the target LLMs, enhancing the computational parallelism of LLMs inference.
In practical applications, lossless speculative decoding allows the target LLMs to generate multiple tokens per step, at the cost of introducing a slight time overhead for each step \cite{leviathan2023fast}.
Distinct from lossy large model inference acceleration techniques (e.g., quantization, pruning), lossless speculative decoding achieves a lossless output through the parallel verification phase of target LLMs \cite{li2024eagle}.
In addition, lossless speculative decoding can be used simultaneously with these widely applied inference acceleration methods because its underlying principle does not conflict with these methods \cite{sun2024triforce}.

The acceleration extent of lossless speculative sampling on target LLMs depends on the time overhead of the draft phase and the accuracy of its output candidates.
The natural approach \cite{chen2023accelerating} that involves using a lower-parameter version from the same LLM series as the draft model suffers from the drawback of excessive time overhead and is unable to accelerate the smallest model in this LLM series.
Several methods have been developed from the perspective of time overhead that do not introduce new models during the draft phase.
Prompt Lookup Decoding (PLD) \cite{saxena2023prompt} matches the last few tokens to somewhere earlier in the input prompt and selects text spans as candidates.
Lookahead \cite{fu2024break} introduces multiple special tokens at the end of the input prompt to enable parallel drafting and transforms the drafts into n-gram candidates.
These methods can consistently enhance the inference speed of the target LLMs due to the low time overhead during the drafting phase.
However, their effectiveness is constrained by the lower accuracy of the output candidates.
To improve inference accuracy, some other studies introduce lightweight draft models to predict candidates.
Medusa \cite{cai2024medusa} utilizes Multi-Layer Perceptron (MLP) as a parallel draft model, which predicts candidates in parallel based on the features (hidden state of the second-to-top layer) of the target LLMs.
Hydra \cite{ankner2024hydra} constructs an autoregressive draft model, employing a transformer decoder layer to compress the prompt into the size of a single feature, serving as the intermediate state for a Recurrent Neural Network (RNN).
Hydra attains a greater level of acceleration compared to Medusa due to its higher accuracy, even though it incurs a higher time overhead because of autoregression.
This indicates that the accuracy in the draft stage has a more significant impact on the overall acceleration extent than the time overhead, when the computational load of the draft model is at the level of a single transformer decoder layer.
EAGLE \cite{li2024eagle1} utilizes a linear combination of the features (hidden state of the second-to-top layer) from target LLMs and the token embeddings as the input for the draft model, and incorporates feature-level loss during training for knowledge distillation.
Notably, as the current state-of-the-art method, EAGLE's draft model is a single unmodified transformer decoder layer, despite various studies \cite{ankner2024hydra,li2024amphista} proposing new draft model structures.
Additionally, EAGLE-2 \cite{li2024eagle}, as an upgraded version of EAGLE, introduces a method for constructing dynamic candidates, further enhancing the performance of EAGLE.

In this paper, we focus on constructing input sequences of the draft model and leveraging the features of the target LLM for knowledge distillation.
We propose FSPAD (Feature Sampling and Partial Alignment Distillation for Lossless Speculative Decoding), which introduces Feature Sampling and Partial Alignment Distillation within the EAGLE-2 framework to boost lossless speculative decoding, based on the following two observations.

\begin{figure}[t]
\centering
\includegraphics[width=1.0\columnwidth]{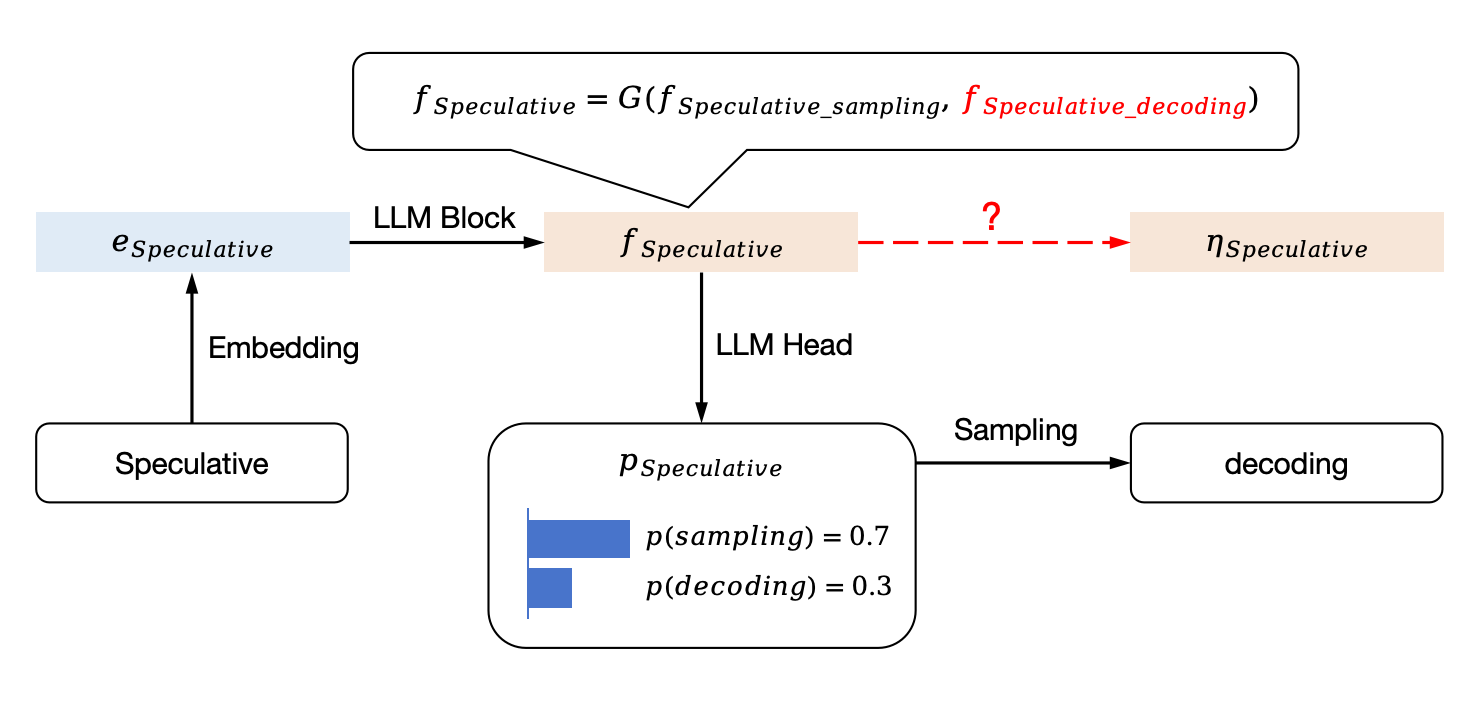} 
\caption{The challenge of addressing the inherent uncertainty while preserving the regular pattern of the feature sequence. Different token components on varying elements in $p_{Speculative}$. However, for feature $f_{Speculative}$, the situation becomes more complex.}
\label{fig2}
\end{figure}

\textbf{The linear combination of features and their sampled results is insufficient to address the inherent uncertainty of the features.}
In this paper, "feature" refers to the hidden state of the second-to-top layer positioned just before the LLM head in the target LLM.
The feature sequence of target LLM is considered to be more regular than the token embedding sequence, therefore, features are utilized as the input for the draft model.
In text generation, the target LLM predicts token distributions and samples based on these predictions, introducing uncertainty.
As shown in Figure \ref{fig2}, the regularity of features arises from the distribution of tokens contained within them, since the distribution of tokens within features is the sole information gain from token embeddings to features.
In an ideal scenario, we aim to mark the tokens of the target LLMs' sampling results while preserving the token distribution within the feature.
However, features are high-dimensional and continuous, and the number of token categories they need to represent is generally much greater than the dimensionality of the features themselves.
For example, in the case of Vicuna 7b, features with a dimensionality of 4096 represent a vocabulary of 32,000 dimensions.
Linear combinations of features and their sampled results clearly cannot address the inherent uncertainty while preserving the regular pattern of the feature sequence.

\begin{figure}[t]
\centering
\includegraphics[width=1.0\columnwidth]{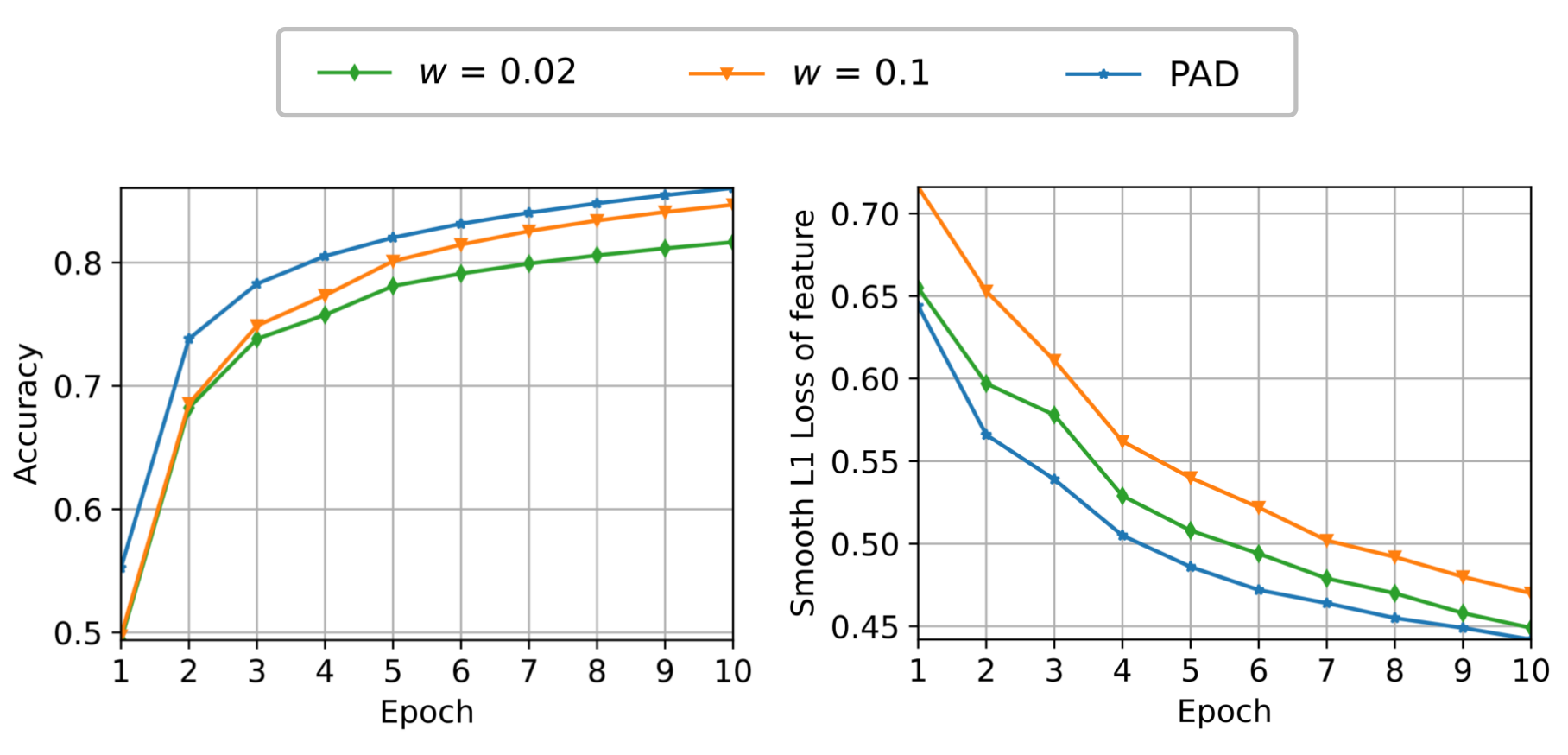} 
\caption{Accuracy and feature-level loss during the training process, where $w$ represents the coefficient of the logit-level loss, and PAD stands for Partial Alignment Distillation in FSPAD.}
\label{fig3}
\end{figure}

\textbf{In the training of a lightweight draft model, there exists a conflict between the feature-level and the logit-level losses.}
The feature-level loss is introduced to facilitate knowledge distillation.
However, to the best of our knowledge, there is currently no research indicating that the knowledge of an LLM can be distilled into a single transformer decoder layer.
We believe that performing strict knowledge distillation between the target LLM and a lightweight draft model is unrealistic.
As shown in Figure \ref{fig3}, we observed a conflict between the feature-level and logit-level losses during the training of the draft model.
We first modify the coefficient $w$ of the logit-level loss in the joint loss during the training of EAGLE.
When $w$ is reduced from 0.1 to 0.02, we observe a significant decrease in feature-level loss, but the prediction accuracy during training decreases.
In Figure \ref{fig3}, we further illustrate the training process after weakening the feature and logit correlation using Partial Alignment Distillation in FSPAD.
Weakening the connection between features and logits can reduce feature-level loss while improving prediction accuracy during training.

Our experiments include multi-turn conversation, translation, summarization, question answering, mathematical reasoning, and retrieval-augmented generation, sourced from the datasets MT-bench \cite{zheng2024judging}, WMT14 \cite{bojar2014findings}, CNN/Daily Mail \cite{nallapati2016abstractive}, Natural Questions \cite{kwiatkowski2019natural}, GSM8K \cite{cobbe2021training}, and DPR \cite{karpukhin2020dense}, respectively.
We select the largest and smallest models from the Vicuna \cite{vicuna} and LLaMA3-Instruct \cite{llama3} series, specifically Vicuna 7B, Vicuna 33B, LLaMA3-Instruct 8B, and LLaMA3-Instruct 70B, as the target LLMs.
Figure \ref{page1} illustrates the performance of FSPAD on the greedy decoding of Vicuna 33B.
For the mathematical reasoning task, FSPAD enables Vicuna 33B to generate 5.3 tokens per step.
We also validated state-of-the-art methods such as EAGLE-2 and other aforementioned studies, conducting a fair comparison with FSPAD.
FSPAD allowed the target model to infer an additional 0.28-0.48 tokens per step compared to EAGLE-2, while following all baseline parameter settings within the framework of EAGLE-2.

FSPAD adds just 0.18B to 1.10B extra parameters for target models whose parameter sizes range from 7B to 70B.
Firstly, these extra parameters do not result in unacceptable training overhead.
We utilize four NVIDIA A100 80G GPUs for training draft models.
For the largest target LLM, LLaMA3-Instruct 70B, in our experiments, the training time for the draft model is within two days.
Secondly, these extra parameters remain small relative to the target LLM's total parameter count.
The time overhead introduced during inference can be offset by a significant improvement in accuracy.

In summary, FSPAD offers the following advantages:

\begin{itemize}
\item FSPAD consistently outperforms state-of-the-art techniques across all tasks and target LLMs, demonstrating robust and stable performance. This is evidenced by the evaluation on multi-turn conversation, translation, summarization, question answering, mathematical reasoning, and retrieval-augmented generation with the Vicuna and LLaMA3-Instruct model series.
\item The components introduced by FSPAD are lightweight and independent of the draft model. This characteristic enhances the adaptability of the FSPAD approach for future research endeavors.
\end{itemize}

\begin{figure}[t]
\centering
\includegraphics[width=1.0\columnwidth]{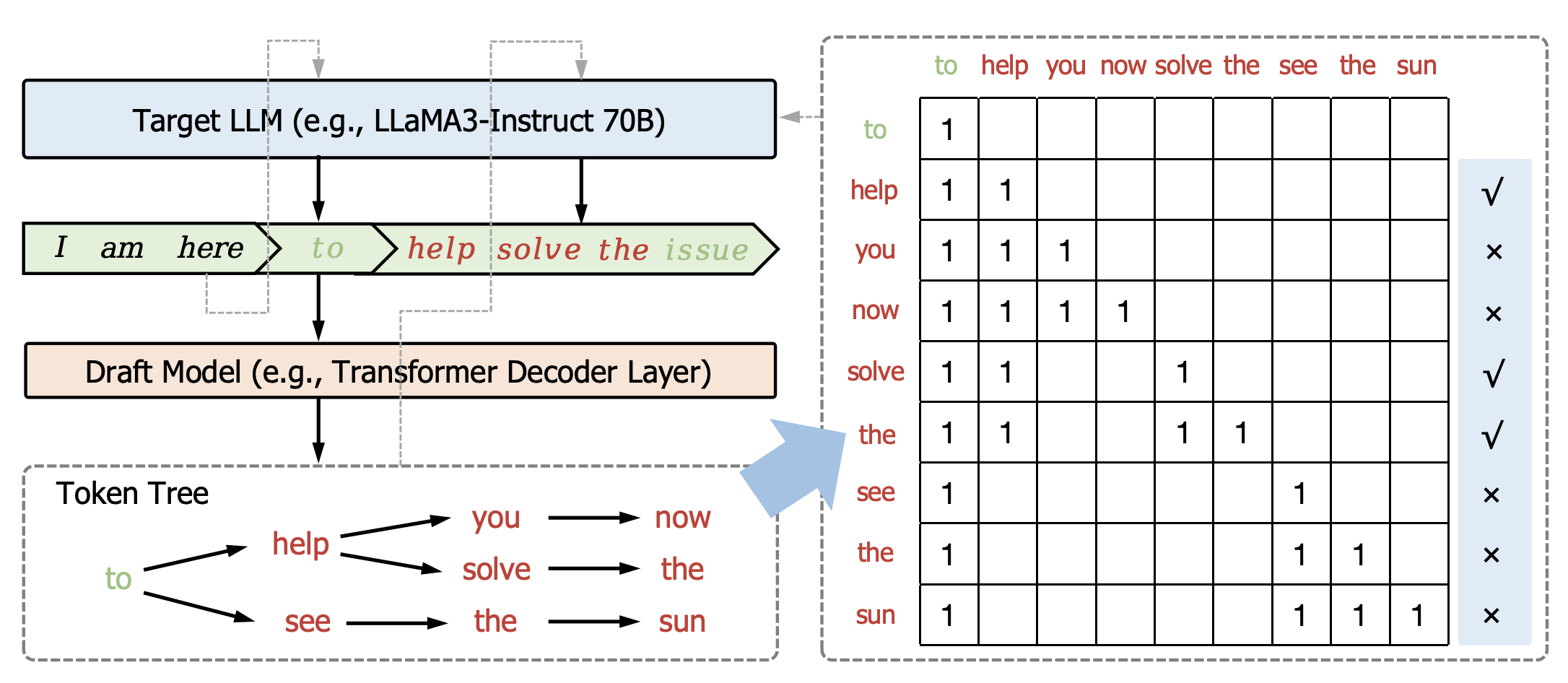} 
\caption{Overview of draft model based speculative decoding.}
\label{background}
\end{figure}

\section{Preliminaries}
Speculative decoding divides the inference of a target LLM into low-cost draft phases and parallel verification phases of the draft results by the target LLM.
Figure \ref{background} illustrates the draft stage and verification stage of a step in draft model based speculative decoding.
This section provides a detailed explanation of the principles behind the draft phase and the verification phase, and also outlines some additional details of our speculative inference framework used in experiments.

In the draft phase, speculative decoding introduces a lightweight draft model or rules with minimal time overhead.
These use the output from the previous step of the target LLM as input to predict the output tokens of the target LLM at multiple subsequent positions.
As mentioned in the previous section, the common practice now is to combine the feature output by the target LLM of the previous step with the token embedding as input to the draft model to utilize as much information as possible.
The methods of predicting these tokens in a draft model may vary.
Medusa predicts tokens at multiple positions simultaneously using multiple heads in parallel.
EAGLE employs a single transformer layer to autoregressively predict these tokens, as this allows leveraging the sequential information among them.
Before these tokens are verified for correctness, two additional steps are required.
Firstly, these tokens need to be organized into a token tree according to their dependencies.
Secondly, the attention mask for the target LLM's next inference step needs to be updated based on this token tree to ensure that the tokens are verified according to their dependencies within the tree.

In the validation phase, the target LLM takes a flattened token tree as input and uses the attention mask constructed during the draft phase to introduce dependencies between the tokens.
This allows the target LLM to output the correct result of the token tree after a single-step inference process.
For lossless speculative decoding, a strict comparison is made between the token tree and these correct results, with differing tokens and their descendant tokens being discarded.
Some research explores non-strict validation strategies, which form lossy speculative decoding.
We do not employ lossy speculative decoding, as it may reduce the effectiveness of text generation by the target LLM.


Another detail within the speculative sampling framework is how the outputs from the draft model are organized into a token tree.
Ideally, the token tree should simultaneously achieve a high step hit rate and a minimal number of tokens, as an excessive number of tokens can lead to a computational bottleneck.
Here, we employ the method used in EAGLE-2 to build a dynamic token tree.
For the autoregressive draft model, the output of each step can be truncated by selecting the top-k.
The dependencies between tokens can be naturally obtained, allowing for the computation of joint probabilities for each token during inference.
Consequently, by applying a top-k operation to all tokens after the draft model inference, we can effectively construct a valid token tree while limiting the total number of tokens.

\begin{figure}[t]
\centering
\includegraphics[width=1.0\columnwidth]{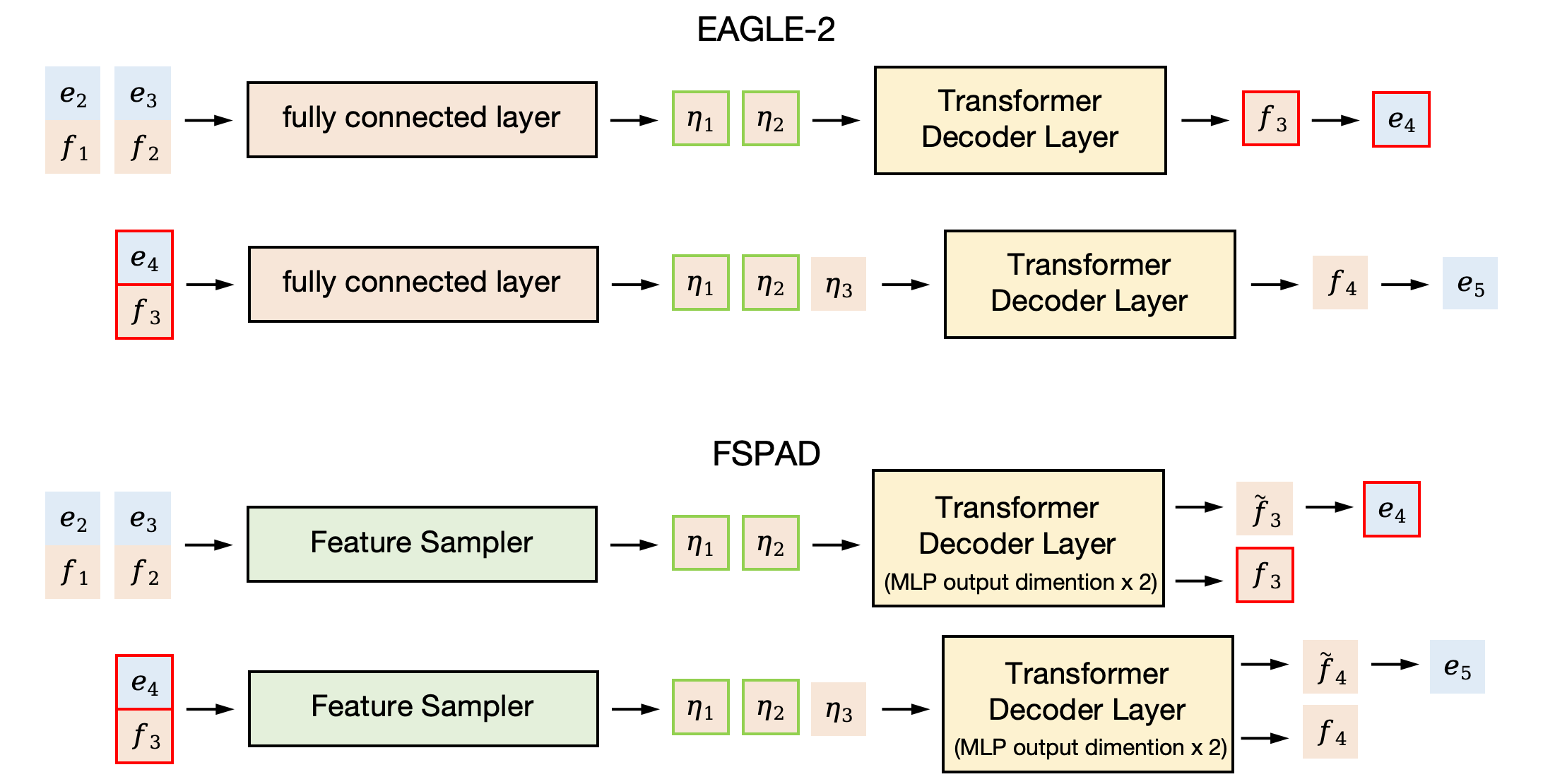} 
\caption{Schematic representation of the drafting phase for EAGLE-2 and FSPAD. $e$ denotes token embeddings, $f$ signifies the features, and $\eta$ represents the inputs of the draft model, with subscripts indicating their positions in the sequence. The red border indicates the predictions of the draft model used for the next step. The green border indicates the inputs of the draft model for the next step.}
\label{structure0}
\end{figure}

\begin{figure}[t]
\centering
\includegraphics[width=1.0\columnwidth]{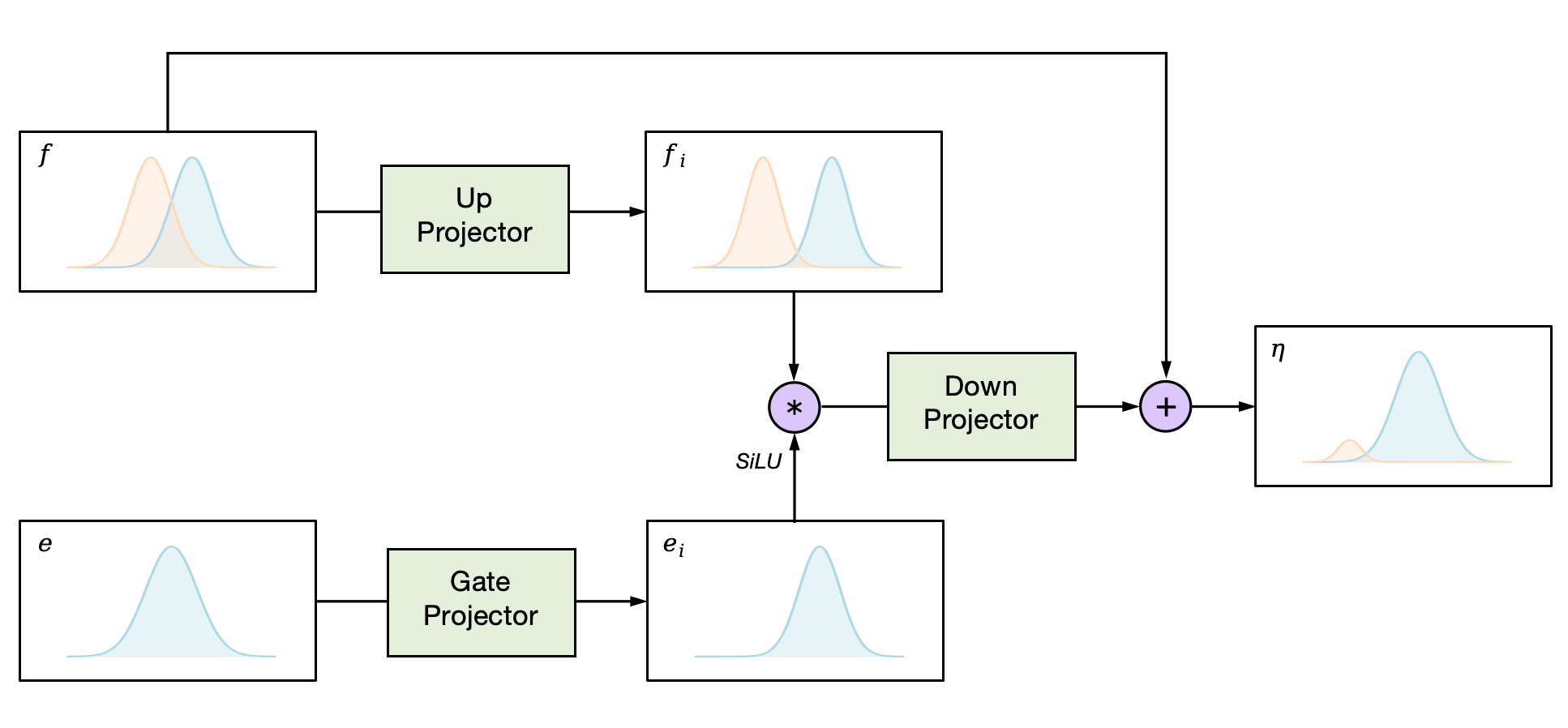} 
\caption{Schematic diagram of the Feature Sampler. The orange and blue shading represents the distribution of different token components in the current space. The subscript $i$ indicates intermediate, and the dimension typically equals the intermediate\_size of the target LLM.}
\label{featuresampler}
\end{figure}

\section{FSPAD}
FSPAD introduces Feature Sampling and Partial Alignment Distillation within the framework of EAGLE-2 to boost speculative decoding.
Figure \ref{structure0} presents a schematic representation of the drafting phase for EAGLE-2 and FSPAD.
Both consist of a connector, depicted on the left half of the figure, and a draft model, shown on the right half.
In our experiments, the draft model is implemented as a Transformer Decoder Layer.
EAGLE-2 and FSPAD synthesize the draft model's input sequence ($\eta_1$, $\eta_2$) using the feature sequence ($f_1$, $f_2$) and the token embedding sequence ($e_2$, $e_3$), advanced by one time step.
The draft model uses input sequence ($\eta_1$, $\eta_2$) to predict $f_3$ and $e_4$,  which are then utilized to synthesize $\eta_3$.
Subsequently, $\eta_3$ is concatenated into the input sequence to predict $f_4$ and $e_5$.
EAGLE-2 employs a fully connected layer as the connector to map the concatenation of $f$ and $e$ into $\eta$, where e is derived from $f$ through the LLM head, sampling process, and embedding layer.
The Feature Sampling of FSPAD replaces the connector with Feature Sampler to obtain $\eta$ that is more suitable for draft model prediction.
The Partial Alignment Distillation in FSPAD adjusts the draft model to produce two features, $\smash{\widetilde{f}}$ and $f$, where $\smash{\widetilde{f}}$ is mapped to $e$ and $f$ is used as input feature for the feature sampler in the subsequent step.

\subsection{Feature Sampling}
As illustrated in Figure \ref{featuresampler}, Feature Sampler of FSPAD consists of three projectors and a sampling operation in a high-dimensional space.
The Feature Sampler takes sequences of $f$ and $e$ with shapes of (bs, seq\_len, hidden\_size), and outputs sequences of $\eta$ with a shape of (bs, seq\_len, hidden\_size).
The Up Projector and Gate Projector linearly map $f$ and $e$ to $f_i$ and $e_i$ with shapes of (bs, seq\_len, intermediate\_size).
The distribution diagram of $f_i$ and $e_i$ in Figure \ref{featuresampler} illustrates the purpose of using the Up Projector and Gate Projector: 
1) transform elements in the feature space of $f$ that are distributed closely together into elements in the feature space of $f_i$ that are distributed further apart by increasing the dimensionality of the feature space; 
2) align the distribution of $e$ with the distribution of the corresponding elements in $f_i$.
The sampling operation in this high-dimensional space is then introduced to label the elements corresponding to $e\_i$.
This process is accomplished by activating $e\_i$ with the SiLU function followed by computing its dot product with $h\_i$.
The Down Projector subsequently maps the sampled sequence back to the shape of the input features (bs, seq\_len, hidden\_size).
This component not only preserves the feature scale of the draft model but also allows for the integration of residual connections.

\subsection{Partial Alignment Distillation}
As illustrated in Figure \ref{structure0}, aligning the features output by the draft model to the target LLM's features and inputting this into the LLM head results in logits that are also aligned with the target LLM's logits.
However, this process undermines the consistent increase in the draft model logits' confidence during training, as determined by the cross-entropy loss function.
The Partial Alignment Distillation in FSPAD doubles the output dimension of the MLP in the transformer decoder layer and shares the same residual connection to achieve output $\smash{\widetilde{f}}$ and $f$.
Partial Alignment Distillation enhances the performance of the draft model with minimal additional computational overhead.
As illustrated by the losses of feature and logit in Figure \ref{structure0}, the approach of mitigating the mutual influence between feature and logit significantly enhances their convergence during the training process.
Moreover, Partial Alignment Distillation does not lead to overfitting, as demonstrated by ablation study experiments.

\subsection{Training of FSPAD}
The task of predicting the next token in LLMs' supervised fine-tuning provides a natural autoregressive task and training dataset for training FSPAD.
Since Feature Sampling and Partial Alignment Distillation only adjust the model architecture and do not have additional outputs that require loss computation, there is no need for further adjustments to the training loss.
For the prediction of the next token, cross-entropy loss is utilized on the dialogue response text to directly optimize this ultimate objective:
\begin{eqnarray}
    p_{n+2}&=&\mathrm{ShiftMask}\left(\mathrm{LLMhead}\left(f_{n+1}\right)\right)\\
    \smash{\hat{p}}_{n+2}&=&\mathrm{ShiftMask}\left(\mathrm{LLMhead}\left(\smash{\hat{f}}_{n+1}\right)\right)\\
    L_{t}&=&\mathrm{CrossEntropy}\left(p_{n+2},\smash{\hat{p}}_{n+2}\right)
\end{eqnarray}
where, ${f}_{n+1}$ denotes the feature output from a single-step inference of the draft model, $\smash{\hat{f}}_{n+1}$ represents the feature output from a single-step inference at the corresponding position of the target LLM, and $\mathrm{ShiftMask}$ refers to the process of shifting the sequence output by the LLM head and masking the question portions within the dialogue training data.
For feature alignment, the Smooth L1 loss is employed to enhance the stability of feature $f$ in draft mode during the autoregressive process:
\begin{eqnarray}
L_{f}&=&\mathrm{SmoothL1}\left(\mathrm{ShiftMask}\left(f_{n+1},\smash{\hat{f}}_{n+1}\right)\right)
\end{eqnarray}
By synthesizing the token prediction loss with the feature alignment loss, we train both the Feature Sampler and the draft model using a composite loss function defined as $L = w\ L_t + L_f$.
We set $w$ to 0.1, consistent with the EAGLE-2.

\section{Experiments}

\begin{table*}[t]
\centering
\begin{tabular}{cccccccccccccc}
\toprule  
 \multicolumn{2}{c}{ } & \multicolumn{2}{c}{MT-bench} & \multicolumn{2}{c}{WMT14} & \multicolumn{2}{c}{CNN/DM} & \multicolumn{2}{c}{Natural Ques.} & \multicolumn{2}{c}{GSM8K} & \multicolumn{2}{c}{DPR} \\
\hline
model & method & $\tau$ & SR & $\tau$ & SR & $\tau$ & SR & $\tau$ & SR & $\tau$ & SR & $\tau$ & SR \\
\hline
\multicolumn{14}{c}{Temperature = 0}\\
\hline
\rule{0pt}{10pt}
\multirow{5}*{V 7B}& PLD & 1.63 & 1.60x & 1.12 & 1.03x & 2.73 & 2.69x & 1.28 & 1.16x & 1.81 & 1.69x & 2.05 & 1.90x \\
\rule{0pt}{5pt}
& Lookahead & 2.20 & 1.68x & 1.91 & 1.10x & 1.65 & 1.41x & 2.10 & 1.46x & 2.78 & 1.97x & 2.15 & 1.44x \\
\rule{0pt}{5pt}
& Medusa & 2.42 & 2.36x & 2.15 & 2.03x & 1.97 & 1.92x & 2.02 & 2.00x & 2.50 & 2.42x & 2.06 & 1.83x \\
\rule{0pt}{5pt}
& EAGLE-2 & 4.74 & 3.56x & 3.92 & 2.55x & 4.27 & 2.99x & 3.62 & 2.68x & 5.03 & 3.69x & 4.49 & 3.31x \\
\rule{0pt}{5pt}
& FSPAD & \textbf{5.12} & \textbf{3.71x} & \textbf{4.05} & \textbf{2.58x} & \textbf{4.60} & \textbf{3.22x} & \textbf{3.93} & \textbf{2.81x} & \textbf{5.32} & \textbf{3.82x} & \textbf{4.79} & \textbf{3.48x} \\
\hline
\rule{0pt}{10pt}
\multirow{5}*{V 33B}& PLD & 1.54 & 1.51x & 1.10 & 1.04x & 2.14 & 2.07x & 1.18 & 1.14x & 1.73 & 1.69x & 1.56 & 1.44x \\
\rule{0pt}{5pt}
& Lookahead & 1.65 & 1.34x & 1.09 & 1.04x & 1.23 & 1.05x & 1.30 & 1.14x & 1.92 & 1.54x & 1.57 & 1.15x \\
\rule{0pt}{5pt}
& Medusa & 2.43 & 2.31x & 2.20 & 2.01x & 2.05 & 1.82x & 2.06 & 2.03x & 2.61 & 2.54x & 2.13 & 1.83x \\
\rule{0pt}{5pt}
& EAGLE-2 & 4.34 & 3.69x & 3.66 & 2.82x & 3.89 & 3.12x & 3.33 & 2.78x & 4.85 & 4.08x & 3.74 & 3.26x \\
\rule{0pt}{5pt}
& FSPAD & \textbf{4.75} & \textbf{3.80x} & \textbf{3.94} & \textbf{2.87x} & \textbf{4.37} & \textbf{3.37x} & \textbf{3.67} & \textbf{3.01x} & \textbf{5.28} & \textbf{4.24x} & \textbf{4.13} & \textbf{3.47x} \\
\hline
\rule{0pt}{10pt}
\multirow{5}*{L3 8B}& PLD & 1.37 & 1.32x & 1.24 & 1.12x & 1.58 & 1.46x & 1.14 & 1.04x & 1.44 & 1.34x & 1.59 & 1.51x \\
\rule{0pt}{5pt}
& Lookahead & 1.49 & 1.35x & 1.35 & 1.19x & 1.42 & 1.26x & 1.32 & 1.17x & 1.71 & 1.45x & 1.52 & 1.26x \\
\rule{0pt}{5pt}
& Medusa & 1.73 & 1.68x & 1.61 & 1.50x & 1.79 & 1.74x & 2.01 & 1.98x & 2.42 & 2.35x & 2.05 & 1.81x \\
\rule{0pt}{5pt}
& EAGLE-2 & 4.21 & 3.36x & 3.87 & 2.97x & 3.94 & 3.02x & 3.63 & 2.74x & 4.52 & 3.51x & 4.32 & 3.42x \\
\rule{0pt}{5pt}
& FSPAD & \textbf{4.55} & \textbf{3.43x} & \textbf{4.11} & \textbf{3.01x} & \textbf{4.16} & \textbf{3.04x} & \textbf{3.84} & \textbf{2.78x} & \textbf{4.82} & \textbf{3.57x} & \textbf{4.67} & \textbf{3.56x} \\
\hline
\rule{0pt}{10pt}
\multirow{5}*{L3 70B}& PLD & 1.34 & 1.33x & 1.15 & 1.10x & 1.54 & 1.48x & 1.14 & 1.08x & 1.42 & 1.37x & 1.54 & 1.47x \\
\rule{0pt}{5pt}
& Lookahead & 1.51 & 1.36x & 1.59 & 1.34x & 1.63 & 1.34x & 1.47 & 1.23x & 1.75 & 1.56x & 1.61 & 1.32x \\
\rule{0pt}{5pt}
& Medusa & 1.79 & 1.75x & 1.68 & 1.65x & 1.82 & 1.78x & 2.03 & 2.01x & 2.47 & 2.36x & 2.06 & 1.98x \\
\rule{0pt}{5pt}
& EAGLE-2 & 4.11 & 3.09x & 3.82 & 2.89x & 3.81 & 2.72x & 3.46 & 2.54x & 4.26 & 3.08x & 4.02 & 2.95x \\
\rule{0pt}{5pt}
& FSPAD & \textbf{4.36} & \textbf{3.14x} & \textbf{4.27} & \textbf{3.03x} & \textbf{3.98} & \textbf{2.74x} & \textbf{3.67} & \textbf{2.61x} & \textbf{4.83} & \textbf{3.34x} & \textbf{4.31} & \textbf{3.00x} \\
\hline
\multicolumn{14}{c}{Temperature = 1}\\
\hline
\rule{0pt}{10pt}
\multirow{2}*{V 7B}& EAGLE-2 & 4.20 & 3.15x & 3.75 & 2.43x & 4.01 & 2.80x & 3.38 & 2.50x & 4.65 & 3.41x & 4.18 & 3.08x \\
\rule{0pt}{5pt}
& FSPAD & \textbf{4.51} & \textbf{3.26x} & \textbf{3.89} & \textbf{2.47x} & \textbf{4.24} & \textbf{2.96x} & \textbf{3.71} & \textbf{2.65x} & \textbf{4.95} & \textbf{3.55x} & \textbf{4.31} & \textbf{3.13x} \\
\hline
\rule{0pt}{10pt}
\multirow{2}*{V 33B}& EAGLE-2 & 4.01 & 3.41x & 3.60 & 2.73x & 3.72 & 2.93x & 3.21 & 2.62x & 4.57 & 3.75x & 3.54 & 3.11x \\
\rule{0pt}{5pt}
& FSPAD & \textbf{4.30} & \textbf{3.46x} & \textbf{3.81} & \textbf{2.77x} & \textbf{4.11} & \textbf{3.15x} & \textbf{3.50} & \textbf{2.79x} & \textbf{4.97} & \textbf{3.90x} & \textbf{4.01} & \textbf{3.27x} \\
\hline
\rule{0pt}{10pt}
\multirow{2}*{L3 8B}& EAGLE-2 & 3.81 & 2.72x & 3.68 & 2.53x & 3.69 & 2.54x & 3.40 & 2.27x & 4.35 & 3.02x & 4.08 & 2.89x \\
\rule{0pt}{5pt}
& FSPAD & \textbf{4.14} & \textbf{2.90x} & \textbf{4.05} & \textbf{2.69x} & \textbf{3.86} & \textbf{2.59x} & \textbf{3.54} & \textbf{2.29x} & \textbf{4.64} & \textbf{3.07x} & \textbf{4.42} & \textbf{3.00x} \\
\hline
\rule{0pt}{10pt}
\multirow{2}*{L3 70B}& EAGLE-2 & 3.72 & 2.50x & 3.63 & 2.46x & 3.56 & 2.28x & 3.24 & 2.10x & 4.09 & 2.64x & 3.79 & 2.57x \\
\rule{0pt}{5pt}
& FSPAD & \textbf{3.96} & \textbf{2.65x} & \textbf{4.20} & \textbf{2.70x} & \textbf{3.69} & \textbf{2.33x} & \textbf{3.38} & \textbf{2.14x} & \textbf{4.65} & \textbf{2.87x} & \textbf{4.07} & \textbf{2.61x} \\
\bottomrule 
\end{tabular}
\caption{Average acceptance lengths ($\tau$) and speedup ratios (SR) of different methods. V represents Vicuna, L3 represents LLaMA3-Instruct. Medusa, PLD, and Lookahead relax acceptance conditions under non-greedy settings, which do not guarantee lossless acceleration. Therefore, we do not compare FSPAD with these methods.}
\label{table1}
\end{table*}

\subsubsection{Models and Tasks.}
To evaluate the effectiveness of FSPAD in the inference of large language models, we conduct a series of experiments utilizing four distinct target models across six tasks. 
We test FSPAD on the smallest and largest models from the Vicuna \shortcite{vicuna} (7B, 33B) and LLaMA3-Instruct \shortcite{llama3} (8B, 70B) series to evaluate its acceleration capabilities across different sizes and types of
models.
In addition, we utilize Spec-Bench \cite{xia2024unlocking} as our benchmark suite to evaluate our model's performance across diverse scenarios.
Spec-Bench encompasses six distinct subtasks: multi-turn conversation, translation, summarization, question answering, mathematical reasoning, and retrieval-augmented generation. 
This benchmark is constructed by randomly selecting 80 instances from each of six widely used datasets: MT-bench \cite{zheng2024judging}, WMT14 \cite{bojar2014findings}, CNN/Daily Mail \cite{nallapati2016abstractive}, Natural Questions \cite{kwiatkowski2019natural}, GSM8K \cite{cobbe2021training}, and DPR \cite{karpukhin2020dense}.
Greedy sampling (temperature=0) and non-greedy sampling (temperature=1) are considered in all experiments for a comprehensive evaluation of speculative decoding performance.
All evaluations are conducted on an NVIDIA A100 80G, except for the 70B model, which utilizes two NVIDIA A100 80G.
\subsubsection{Metrics.}
FSPAD neither has relaxed acceptance conditions nor requires fine-tuning of the target model, thereby achieving lossless inference acceleration.
Therefore, to assess the acceleration performance of FSPAD on target LLMs, we utilize two main metrics: average acceptance length ($\tau$) and speedup ratio (SR).
The average acceptance length measures the average number of tokens accepted per forward pass by the target large language models, excluding any overhead of retrieving or constructing draft tokens, indicating the maximum possible acceleration.
The second metric is the actual speedup ratio relative to vanilla autoregressive decoding.
\subsubsection{Baseline.}
In this study, we only investigate lossless speculative decoding approaches of LLMs.
In approaches that do not rely on draft models, we examine Prompt Lookup Decoding \cite{saxena2023prompt} (PLD) and Lookahead Decoding \cite{fu2024break}, both of which have already been integrated into the popular inference framework vLLM \cite{kwon2023efficient}.
In methods utilizing draft models, we examine Medusa \cite{cai2024medusa} and EAGLE-2 \cite{li2024eagle}, which represent parallel speculative decoding and autoregressive speculative decoding in feature-level speculative decoding, respectively.
Additionally, EAGLE-2 is considered the state-of-the-art method for lossless speculative decoding tasks.
Collectively, these baseline methods provide a solid framework for evaluating the efficiency of FSPAD in the LLM decoding process.
\subsubsection{Training.}
We use the SharedGPT dataset \shortcite{sharedgpt}, which includes 68,000 dialogues from the Vicuna series models' supervised fine-tuning dataset, as our training corpus.
Due to the significant time and computational resources required, we opt not to regenerate responses for each dialogue turn using the target LLMs.
Conducting training without regenerated data across all comparative methods remains equitable, although previous work \cite{li2024eagle1} indicates that such an approach could slightly enhance the performance of the draft model.
The learning rate is set to 5e-5, with  $(\beta_1=0.9,\beta_2=0.95)$ for the AdamW optimizer and implemented gradient clipping of 0.5.
The training parameters for FSPAD are 0.42B, 0.42B, 1.09B, and 2.05B, corresponding to the target LLMs with parameter sizes of 7B, 8B, 33B, and 70B, respectively.
We utilized four NVIDIA A100 80G GPUs for the training process.

\subsection{Effectiveness of FSPAD}
Table \ref{table1} presents the average acceptance lengths ($\tau$) and speedup ratios (SR) of different methods.
The characteristics of each method can be discerned from Table \ref{table1}.
PLD's capability on the Vicuna summarization task (CNN/DM) surpasses its performance in other tasks, due to PLD's retrieval-based draft generation and the high overlap in context when Vicuna performs summarization.
By keeping track of the trajectory of Jacobi decoding, Lookahead outperforms PLD in all tasks except for summarization.
Medusa achieves a speedup ratio similar to its average acceptance length, which is better than methods not requiring draft models, thanks to Medusa's high parallelism and lightweight draft model.
FSPAD and EAGLE-2, due to their autoregressive draft models and higher computational complexity, achieved 2-3 times the average acceptance length of the aforementioned methods, at the cost of significantly lower speedup ratios relative to their average acceptance length.
All these methods underperformed on the LLaMA3-Instruct series models compared to the Vicuna series models.
This is because SharedGPT (i.e., the training data used in our experiments) is the SFT dataset for the Vicuna series models, whereas the SFT dataset for the LLaMA3-Instruct series models is not open-source.

Across all tasks and large language models (LLMs) we tested, FSPAD achieved the highest values in both average acceptance lengths and speedup ratios.
EAGLE-2 is the method with performance closest to FSPAD among these comparison methods.
In translation tasks (WMT14) where both input and output texts are relatively short, its performance is comparable to FSPAD.
Nevertheless, in various other tasks, the performance of FSPAD is markedly superior to that of EAGLE-2.
We illustrate the advantages of FSPAD using the summarization task (CNN/DM), characterized by longer input texts, and the mathematical reasoning task (GSM8K), noted for longer output texts.
For the summarization task (CNN/DM) with Vicuna 33B, at a temperature setting of 0, FSPAD demonstrates a 12.3\% enhancement in average acceptance length and an 8.0\% increase in speedup ratios compared to EAGLE-2.
When the temperature was set to 1, FSPAD still outperforms EAGLE-2 with a 10.4\% improvement in average acceptance length and a 7.5\% enhancement in speedup ratios.
For the mathematical reasoning task (GSM8K) with LLaMA3-Instruct 70B, at a temperature setting of 0, FSPAD demonstrates a 13.3\% enhancement in average acceptance length and an 8.4\% increase in speedup ratios compared to EAGLE-2.
When the temperature was set to 1, FSPAD still outperforms EAGLE-2 with a 13.6\% improvement in average acceptance length and an 8.7\% enhancement in speedup ratios.

\begin{table}[t]
\centering
\begin{tabular}{cccccc}
\toprule  
& & \multicolumn{2}{c}{MT-bench} &  \multicolumn{2}{c}{CNN/DM} \\
\hline
model & method & $\tau$ & SR & $\tau$ & SR \\
\hline
\multicolumn{6}{c}{Temperature = 0}\\
\hline
\rule{0pt}{10pt}
\multirow{2}*{V 7B}&EAGLE-2&4.22&3.44x&3.87&3.21x\\
\rule{0pt}{5pt}
 &FSPAD&\textbf{4.51}&\textbf{3.61x}&\textbf{4.14}&\textbf{3.39x}\\
\hline
\rule{0pt}{10pt}
\multirow{2}*{V 33B}&EAGLE-2&3.91&3.61x&3.54&3.21x\\
\rule{0pt}{5pt}
 &FSPAD&\textbf{4.23}&\textbf{3.74x}&\textbf{3.95}&\textbf{3.41x}\\
 \hline
\multicolumn{6}{c}{Temperature = 1}\\
\hline
\rule{0pt}{10pt}
\multirow{2}*{V 7B}&EAGLE-2&3.67&3.08x&3.75&3.05x\\
\rule{0pt}{5pt}
 &FSPAD&\textbf{3.96}&\textbf{3.16x}&\textbf{3.91}&\textbf{3.07x}\\
\hline
\rule{0pt}{10pt}
\multirow{2}*{V 33B}&EAGLE-2&3.66&3.39x&3.41&3.05x\\
\rule{0pt}{5pt}
 &FSPAD&\textbf{3.97}&\textbf{3.49x}&\textbf{3.86}&\textbf{3.30x}\\
\bottomrule 
\end{tabular}
\caption{Average acceptance lengths ($\tau$) and speedup ratios (SR) on fewer candidate tokens.}
\label{table2}
\end{table}

\subsection{Performance on Fewer Candidate Tokens}
Setting the number of candidate tokens to 60 has produced outstanding experimental results.
However, in practical inference frameworks, constructing such a large number of candidate tokens is typically infeasible due to computational constraints \cite{zhang2024recurrent}.
Consequently, we conduct experiment with a reduced number of candidate tokens to further illustrate the versatility and general applicability of FSPAD. 
Specifically, we half the total number of tokens and the top k during the construction of the token tree.
As shown in Table 2, for multi-turn dialogue tasks (MT-bench), FSPAD demonstrates an average acceptance length gain of 6.8\%-8.4\% and a speedup gain of 2.5\%-4.9\% compared to EAGLE-2.
In summarization tasks (CNN/DM), FSPAD achieves an average acceptance length gain of 4.2\%-13.1\% and a speedup gain of 0.6\%-8.1\%.

\subsection{Ablation Study}
We conduct the ablation study on multi-turn dialogue (MT-bench), summarization (CNN/DM), and mathematical reasoning (GSM8K) tasks using the Vicuna 7B model to individually analyze the impact of Feature Sampling (FS) and Partial Alignment Distillation (PAD) introduced by FSPAD.
\subsubsection{Feature Sampler}
As shown in Figure \ref{fig2}, there is inherent uncertainty in the features of the target LLM.
Directly using these features as input to the model lacks the sampling information of the target model at this step (i.e., the specific token output by this feature).
Therefore, in FSPAD, token embeddings are sampled in high-dimensional space through Feature Sampler.
To validate the effectiveness of the Feature Sampler, we conducted ablation studies using the method of directly linearly combining token embeddings and features from EAGLE-2 as a substitute.
The experimental results in Table \ref{table3} show that the average acceptance length and speedup ratio are both higher across the three tasks when the Feature Sampler is introduced, demonstrating the effectiveness of the Feature Sampler.
\subsubsection{Partial Alignment Distillation}
Figure \ref{fig3} illustrates the antagonism between feature alignment and the confidence of model output logits during strict knowledge distillation.
Therefore, FSPAD reduces the interference between feature-level and logit-level losses during training through Partial Alignment Distillation, while almost not increasing computational overhead.
Partial Alignment Distillation does not require any replacement of components, as it simply increases the dimensionality of the MLP output in the transformer layer.
The experimental results in Table \ref{table3} indicate that the introduction of Partial Alignment Distillation enhances the average acceptance length and speedup ratio across three tasks.
Notably, the improvement is most significant in multi-turn dialogues (MT-bench), which are most similar to the training data from SharedGPT, while other tasks also show gains, indicating that there is no overfitting.
\begin{table}[t]
\centering
\begin{tabular}{ccccccc}
\toprule
& \multicolumn{2}{c}{MT-bench} & \multicolumn{2}{c}{CNN/DM} & \multicolumn{2}{c}{GSM8K} \\
\hline
method&$\tau$&SR&$\tau$&SR&$\tau$&SR\\
\hline
\multicolumn{7}{c}{Temperature = 0}\\
\hline
\rule{0pt}{10pt}
w/o both &4.74&3.56x&4.27&2.99x&5.03&3.69x\\
\rule{0pt}{5pt}
w/o FS &4.88&3.65x&4.30&3.01x&5.04&3.69x\\
\rule{0pt}{5pt}
w/o PAD &5.10&3.69x&4.56&3.19x&5.21&3.74x\\
\rule{0pt}{5pt}
FSPAD &\textbf{5.12}&\textbf{3.71x}&\textbf{4.60}&\textbf{3.22x}&\textbf{5.32}&\textbf{3.82x}\\
 \hline
\multicolumn{7}{c}{Temperature = 1}\\
\hline
\rule{0pt}{10pt}
w/o both &4.20&3.15x&4.01&2.80x&4.65&3.41x\\
\rule{0pt}{5pt}
w/o FS &4.32&3.24x&4.06&2.83x&4.77&3.49x\\
\rule{0pt}{5pt}
w/o PAD &4.44&3.20x&4.18&2.91x&4.89&3.50x\\
\rule{0pt}{5pt}
FSPAD &\textbf{4.51}&\textbf{3.26x}&\textbf{4.24}&\textbf{2.96x}&\textbf{4.95}&\textbf{3.55x}\\
\bottomrule
\end{tabular}
\caption{Ablation Study on Vicuan 7B.}
\label{table3}
\end{table}

\section{Related work}

Lossless speculative decoding \cite{leviathan2023fast} achieves lossless acceleration of LLMs by dividing the inference of each step of the LLMs into a draft stage and a verification stage.
Previous research primarily differs in the model architectures or rules used during the draft stage.
PLD \cite{saxena2023prompt} and Lookahead \cite{fu2024break} retrieve similar segments in the prompt as candidates.
Medusa \cite{cai2024medusa} employs a series of MLPs as a parallel draft model to predict candidates.
Hydra \cite{ankner2024hydra} and Recurrent Drafter \cite{zhang2024recurrent} use an RNN-based draft model for regressive candidate prediction.
EAGLE \cite{li2024eagle1} employs a layer of transformer decoder as the draft model and uses autoregression of feature sequences instead of token-level autoregression.
In contrast, FSPAD primarily focuses on constructing input sequences suitable for lightweight draft model prediction and the training methods for the draft model.

\section{Conclusion}
In this paper, we introduce FSPAD, which introduces Feature Sampling and Partial Alignment Distillation within the existing framework to boost lossless speculative decoding. 
In our study, we conducted a comprehensive evaluation using both greedy and non-greedy decoding strategies on the largest and smallest models from the Vicuna and LLaMA3-Instruct series. The evaluation also covered various tasks of multi-turn conversation, translation, summarization, question answering, mathematical reasoning, and retrieval-augmented generation.
The results show that FSPAD outperforms the state-of-the-art method across all the aforementioned tasks and target LLMs.

\bibliography{aaai25}

\begin{thebibliography}{23}
\providecommand{\natexlab}[1]{#1}

\bibitem[{Achiam et~al.(2023)Achiam, Adler, Agarwal, Ahmad, Akkaya, Aleman, Almeida, Altenschmidt, Altman, Anadkat et~al.}]{achiam2023gpt}
Achiam, J.; Adler, S.; Agarwal, S.; Ahmad, L.; Akkaya, I.; Aleman, F.~L.; Almeida, D.; Altenschmidt, J.; Altman, S.; Anadkat, S.; et~al. 2023.
\newblock Gpt-4 technical report.
\newblock \emph{arXiv preprint arXiv:2303.08774}.

\bibitem[{Ankner et~al.(2024)Ankner, Parthasarathy, Nrusimha, Rinard, Ragan-Kelley, and Brandon}]{ankner2024hydra}
Ankner, Z.; Parthasarathy, R.; Nrusimha, A.; Rinard, C.; Ragan-Kelley, J.; and Brandon, W. 2024.
\newblock Hydra: Sequentially-dependent draft heads for medusa decoding.
\newblock \emph{arXiv preprint arXiv:2402.05109}.

\bibitem[{Bojar et~al.(2014)Bojar, Buck, Federmann, Haddow, Koehn, Leveling, Monz, Pecina, Post, Saint-Amand et~al.}]{bojar2014findings}
Bojar, O.; Buck, C.; Federmann, C.; Haddow, B.; Koehn, P.; Leveling, J.; Monz, C.; Pecina, P.; Post, M.; Saint-Amand, H.; et~al. 2014.
\newblock Findings of the 2014 workshop on statistical machine translation.
\newblock In \emph{Proceedings of the ninth workshop on statistical machine translation}, 12--58.

\bibitem[{Cai et~al.(2024)Cai, Li, Geng, Peng, Lee, Chen, and Dao}]{cai2024medusa}
Cai, T.; Li, Y.; Geng, Z.; Peng, H.; Lee, J.~D.; Chen, D.; and Dao, T. 2024.
\newblock Medusa: Simple llm inference acceleration framework with multiple decoding heads.
\newblock \emph{arXiv preprint arXiv:2401.10774}.

\bibitem[{Chen et~al.(2023)Chen, Borgeaud, Irving, Lespiau, Sifre, and Jumper}]{chen2023accelerating}
Chen, C.; Borgeaud, S.; Irving, G.; Lespiau, J.-B.; Sifre, L.; and Jumper, J. 2023.
\newblock Accelerating large language model decoding with speculative sampling.
\newblock \emph{arXiv preprint arXiv:2302.01318}.

\bibitem[{Cobbe et~al.(2021)Cobbe, Kosaraju, Bavarian, Chen, Jun, Kaiser, Plappert, Tworek, Hilton, Nakano et~al.}]{cobbe2021training}
Cobbe, K.; Kosaraju, V.; Bavarian, M.; Chen, M.; Jun, H.; Kaiser, L.; Plappert, M.; Tworek, J.; Hilton, J.; Nakano, R.; et~al. 2021.
\newblock Training verifiers to solve math word problems.
\newblock \emph{arXiv preprint arXiv:2110.14168}.

\bibitem[{Fu et~al.(2024)Fu, Bailis, Stoica, and Zhang}]{fu2024break}
Fu, Y.; Bailis, P.; Stoica, I.; and Zhang, H. 2024.
\newblock Break the sequential dependency of llm inference using lookahead decoding.
\newblock \emph{arXiv preprint arXiv:2402.02057}.

\bibitem[{Karpukhin et~al.(2020)Karpukhin, O{\u{g}}uz, Min, Lewis, Wu, Edunov, Chen, and Yih}]{karpukhin2020dense}
Karpukhin, V.; O{\u{g}}uz, B.; Min, S.; Lewis, P.; Wu, L.; Edunov, S.; Chen, D.; and Yih, W.-t. 2020.
\newblock Dense passage retrieval for open-domain question answering.
\newblock \emph{arXiv preprint arXiv:2004.04906}.

\bibitem[{Kwiatkowski et~al.(2019)Kwiatkowski, Palomaki, Redfield, Collins, Parikh, Alberti, Epstein, Polosukhin, Devlin, Lee et~al.}]{kwiatkowski2019natural}
Kwiatkowski, T.; Palomaki, J.; Redfield, O.; Collins, M.; Parikh, A.; Alberti, C.; Epstein, D.; Polosukhin, I.; Devlin, J.; Lee, K.; et~al. 2019.
\newblock Natural questions: a benchmark for question answering research.
\newblock \emph{Transactions of the Association for Computational Linguistics}, 7: 453--466.

\bibitem[{Kwon et~al.(2023)Kwon, Li, Zhuang, Sheng, Zheng, Yu, Gonzalez, Zhang, and Stoica}]{kwon2023efficient}
Kwon, W.; Li, Z.; Zhuang, S.; Sheng, Y.; Zheng, L.; Yu, C.~H.; Gonzalez, J.~E.; Zhang, H.; and Stoica, I. 2023.
\newblock Efficient Memory Management for Large Language Model Serving with PagedAttention.
\newblock In \emph{Proceedings of the ACM SIGOPS 29th Symposium on Operating Systems Principles}.

\bibitem[{Leviathan, Kalman, and Matias(2023)}]{leviathan2023fast}
Leviathan, Y.; Kalman, M.; and Matias, Y. 2023.
\newblock Fast inference from transformers via speculative decoding.
\newblock In \emph{International Conference on Machine Learning}, 19274--19286. PMLR.

\bibitem[{Li et~al.(2024{\natexlab{a}})Li, Wei, Zhang, and Zhang}]{li2024eagle}
Li, Y.; Wei, F.; Zhang, C.; and Zhang, H. 2024{\natexlab{a}}.
\newblock EAGLE-2: Faster Inference of Language Models with Dynamic Draft Trees.
\newblock \emph{arXiv preprint arXiv:2406.16858}.

\bibitem[{Li et~al.(2024{\natexlab{b}})Li, Wei, Zhang, and Zhang}]{li2024eagle1}
Li, Y.; Wei, F.; Zhang, C.; and Zhang, H. 2024{\natexlab{b}}.
\newblock Eagle: Speculative sampling requires rethinking feature uncertainty.
\newblock \emph{arXiv preprint arXiv:2401.15077}.

\bibitem[{Li et~al.(2024{\natexlab{c}})Li, Yang, Gao, Liu, Liu, Li, Peng, Tian, and Barsoum}]{li2024amphista}
Li, Z.; Yang, X.; Gao, Z.; Liu, J.; Liu, Z.; Li, D.; Peng, J.; Tian, L.; and Barsoum, E. 2024{\natexlab{c}}.
\newblock Amphista: Accelerate LLM Inference with Bi-directional Multiple Drafting Heads in a Non-autoregressive Style.
\newblock \emph{arXiv preprint arXiv:2406.13170}.

\bibitem[{Meta()}]{llama3}
Meta. 2024.
\newblock {Introducing Meta Llama 3: The most capable openly available LLM to date}.
\newblock \url{https://ai.meta.com/blog/meta-llama-3/}.

\bibitem[{Nallapati et~al.(2016)Nallapati, Zhou, Gulcehre, Xiang et~al.}]{nallapati2016abstractive}
Nallapati, R.; Zhou, B.; Gulcehre, C.; Xiang, B.; et~al. 2016.
\newblock Abstractive text summarization using sequence-to-sequence rnns and beyond.
\newblock \emph{arXiv preprint arXiv:1602.06023}.

\bibitem[{Saxena(2023)}]{saxena2023prompt}
Saxena, A. 2023.
\newblock Prompt Lookup Decoding.

\bibitem[{SharedGPT()}]{sharedgpt}
SharedGPT. 2023.
\newblock {SharedGPT}.
\newblock \url{https://huggingface.co/datasets/Aeala/ShareGPT_Vicuna_unfiltered/}.

\bibitem[{Sun et~al.(2024)Sun, Chen, Yang, Tian, and Chen}]{sun2024triforce}
Sun, H.; Chen, Z.; Yang, X.; Tian, Y.; and Chen, B. 2024.
\newblock Triforce: Lossless acceleration of long sequence generation with hierarchical speculative decoding.
\newblock \emph{arXiv preprint arXiv:2404.11912}.

\bibitem[{Vicuna()}]{vicuna}
Vicuna. 2023.
\newblock {Vicuna: An open-source chatbot impressing gpt-4 with 90\%* chatgpt quality}.
\newblock \url{https://lmsys.org/blog/2023-03-30-vicuna/}.

\bibitem[{Xia et~al.(2024)Xia, Yang, Dong, Wang, Li, Ge, Liu, Li, and Sui}]{xia2024unlocking}
Xia, H.; Yang, Z.; Dong, Q.; Wang, P.; Li, Y.; Ge, T.; Liu, T.; Li, W.; and Sui, Z. 2024.
\newblock Unlocking efficiency in large language model inference: A comprehensive survey of speculative decoding.
\newblock \emph{arXiv preprint arXiv:2401.07851}.

\bibitem[{Zhang et~al.(2024)Zhang, Wang, Wang, Zhang, and Cheng}]{zhang2024recurrent}
Zhang, A.; Wang, C.; Wang, Y.; Zhang, X.; and Cheng, Y. 2024.
\newblock Recurrent drafter for fast speculative decoding in large language models.
\newblock \emph{arXiv preprint arXiv:2403.09919}.

\bibitem[{Zheng et~al.(2024)Zheng, Chiang, Sheng, Zhuang, Wu, Zhuang, Lin, Li, Li, Xing et~al.}]{zheng2024judging}
Zheng, L.; Chiang, W.-L.; Sheng, Y.; Zhuang, S.; Wu, Z.; Zhuang, Y.; Lin, Z.; Li, Z.; Li, D.; Xing, E.; et~al. 2024.
\newblock Judging llm-as-a-judge with mt-bench and chatbot arena.
\newblock \emph{Advances in Neural Information Processing Systems}, 36.

\end{thebibliography}

\end{document}